\documentclass[letterpaper]{article} 
\usepackage{aaai2026}  
\usepackage{times}  
\usepackage{helvet}  
\usepackage{courier}  
\usepackage[hyphens]{url}  
\usepackage{graphicx} 
\usepackage{natbib}  
\usepackage{caption} 
\usepackage{amsmath}
\usepackage{amssymb}
\usepackage{algorithm}
\usepackage{algorithmic}

\usepackage{newfloat}
\usepackage{listings}
\DeclareCaptionStyle{ruled}{labelfont=normalfont,labelsep=colon,strut=off} 
\floatstyle{ruled}
\newfloat{listing}{tb}{lst}{}
\floatname{listing}{Listing}
\title{Improving Fake Profile Detection Through Multimodal Learning and Research-Guided Feature Integration}
\author{
    Jean Claude Hirwa,
    Jamiu Dahunsi,
    M. Giramata Candide,
    Lowami Uwimana,
    Thierry Henry Ishimwe
}
\affiliations{
    \textsuperscript{\rm 1}Carnegie Mellon University Africa\\
    \{jhirwa, jdahunsi, mgiramat, ulowami, ithierry\}@andrew.cmu.edu
}

\usepackage{bibentry}

\begin{document}

\maketitle

\begin{abstract}
Fake accounts continue to undermine online platforms, yet it remains unclear whether multimodal learning offers meaningful improvements over strong metadata baselines or graph-based approaches. In this work, we study bot detection by integrating textual, visual, and account metadata signals on the TwiBot-22 benchmark using a subsample of 30,000 users. We propose a multimodal framework combining pretrained textual (XLM-RoBERTa) and visual (CLIP) embeddings with structured account-level features such as follower counts, posting frequency, and account age, and evaluate multiple modality combinations using CatBoost alongside a Relational Graph Convolutional Network (R-GCN). Results show that metadata alone is highly predictive, while multimodal fusion yields only marginal and largely statistically insignificant gains. The graph-based model underperforms classical variants under the subsampled graph conditions of this study. Overall, our findings suggest that under practical constraints, feature quality and metadata richness matter more than fusion complexity.
\end{abstract}


\section{Introduction}
Fake accounts persist in compromising the integrity of online platforms by propagating misinformation, manipulating public discourse, and enabling large-scale fraudulent activities. As social media ecosystems grow, adversaries are increasingly able to design accounts that mimic legitimate user behavior, making reliable detection significantly more challenging. Traditional detection systems, which often rely on a single modality such as textual content or account metadata, struggle to remain robust against these evolving strategies.

Recent advances in multimodal machine learning suggest that integrating heterogeneous signals such as text, images, and behavioral metadata can improve detection performance by capturing complementary patterns that are difficult to manipulate simultaneously \cite{choudhary_mmds_2024, ArranzEscudero2025EnhancingMCA}. At the same time, graph-based approaches have demonstrated that relational information encoded in social networks provides additional signals about coordinated or anomalous behavior \cite{feng2021botrgcn, gera2022tbot}. Despite these advances, it remains unclear how much multimodal fusion actually improves performance over strong metadata-based baselines.

In this work, we investigate whether combining visual, textual, and metadata features leads to meaningful improvements in fake profile detection, and how these gains compare to those obtained through relational modeling. Specifically, we design a multimodal framework that integrates representations derived from profile images, tweets, and account-level metadata features, and evaluate it alongside a graph-based model that captures user interactions. This dual perspective allows us to contrast instance-level learning with relational modeling under consistent experimental conditions. The study is scoped to publicly available Twitter data via the TwiBot-22 benchmark, chosen for its breadth of multimodal signals and the availability of social network structure.

The study is guided by two central questions. First, does combining visual and textual information with metadata significantly improve bot detection performance? Second, can incorporating social network structure provide additional gains beyond feature-based approaches? To answer these questions, we evaluate multiple modality combinations and modeling paradigms on the TwiBot-22 benchmark, with a focus on understanding the relative contribution of each signal type.

Our contributions are threefold. First, we develop a multimodal detection pipeline that integrates pretrained visual and textual encoders with structured metadata features in a unified representation. Second, we provide a systematic empirical analysis of modality contributions through ablation studies across multiple feature combinations. Third, we compare classical and graph-based modeling approaches under realistic sub-sampled graph conditions to assess the utility of relational information for bot detection.

\section{Related Work}
\subsection{Multimodal and Visual Foundations}
Early approaches to automated account detection relied primarily on handcrafted features derived from account metadata and posting behavior \cite{rossetti_bots_2023}. As these methods proved vulnerable to adversarial mimicry, the research community shifted toward richer signal combinations. Dracewicz and Sepczuk demonstrated that holistic image-based analysis of profile pages can achieve high accuracy, though distinguishing fully autonomous bots from partially autonomous cyborg accounts remained a key challenge \cite{x_fake}. This limitation motivated the integration of complementary modalities, leading to frameworks that combine visual, textual, and statistical signals within unified architectures.

Subsequent work explored increasingly sophisticated fusion strategies. The MSM-BD framework fuses image, text, and statistical data using ResNet-18 with cross-modal residual attention \cite{wu2025msmbd}, while FakedBits pairs EfficientNet-B0 for visual processing with distilBERT for text, employing late fusion at the classification layer to mitigate noise from heterogeneous data streams \cite{sharma2023fakedbits}. Moving beyond modality-level fusion, DMIBot introduces a synergistic interaction module with recursive deep interaction mechanisms to jointly model metadata, text, and graph signals \cite{lin2025dmibot}. Arranz-Escudero et al. (2025) further enriched the feature space by incorporating semantic embeddings and novel relational signals such as list ownership \cite{ArranzEscudero2025EnhancingMCA}. On the classifier side, the CB-MTE framework demonstrates that ensemble methods such as CatBoost can effectively handle the diverse data distributions arising in multi-source detection settings \cite{cheng2025cbmte}.

\subsection{Graph-Based and Relational Models}
The recognition that bot accounts rarely operate in isolation motivated a parallel line of research focused on social network structure. The TwiBot-22 benchmark \cite{twibot-22} formalized this direction by providing large-scale graph data alongside multimodal user features, enabling the analysis of structural relationships at scale. BotRGCN was among the first to exploit this structure through relational graph convolutional networks, modeling different interaction types as distinct edge relations \cite{feng2021botrgcn}. The Relational Graph Transformer (RGT) extended this by introducing attention mechanisms that prioritize high-influence nodes, accounting for heterogeneity in user interactions \cite{feng2022rgt}. Social network analysis approaches have also been proposed as alternatives to feature-based methods, classifying bots based on behaviors observable within the network topology alone \cite{bui2022twitter}, while T-Bot addressed the structural complexity of trend-centric networks by modeling the evolving nature of social interactions \cite{gera2022tbot}.

More recent work has addressed fundamental limitations of standard graph representations. HCGBot constructs a homophilous context graph that captures implicit interactions not visible in explicit follower-following links, improving detection in realistic deployment conditions \cite{wan2026hcgbot}. HHG-Bot integrates hypergraphs and heterogeneous graph techniques to model high-order interaction features that pairwise edges cannot represent \cite{wang2025hhgbot}. To reduce dependence on large annotated datasets, the CBD framework applies graph contrastive learning \cite{zhou2023cbd}, while SRGAT employs semi-supervised graph attention to leverage unlabeled data \cite{huang2025srgat}. RoGBot offers a relationship-oblivious alternative for settings where explicit social links are sparse or unavailable \cite{anshul2025rogbot}, and BotDGT incorporates dynamicity-aware transformers to capture how network structure evolves over time \cite{he2024botdgt}.

\subsection{LLM-Enhanced and Community-Aware Detection}
The rise of large language models has introduced both new capabilities and new challenges for bot detection. Feng et al. (2024) characterize this dual role, noting that while LLMs improve semantic reasoning in detection systems, they also enable adversaries to generate more convincing automated content \cite{feng2024whatbot}. In response, SeGA leverages LLM-generated prompts within a preference-aware self-contrastive learning framework to identify anomalous behavioral patterns \cite{zhang2024sega}, and Sallah et al. (2024) show that fine-tuned transformer classifiers can detect nuanced behaviors that evade traditional rule-based filters \cite{sallah2024finetuned}. At a broader scale, BotMoE addresses the contextual variability of bot behavior across communities by employing a mixture-of-experts architecture that adapts detection strategies to specific social contexts \cite{liu2023botmoe}.

\subsection{Behavioral Patterns and Content Analysis}
Beyond structural and semantic signals, temporal and behavioral consistency provide complementary signatures of automation that are difficult to mimic at scale. Large-scale studies of over 200 million users have identified systematic differences in communication structure and linguistic cue usage between human and automated accounts \cite{ng2025global}, while analyses of major geopolitical events such as the UK 2019 General Election \cite{bruno2022brexit} and the US 2020 electoral cycle \cite{shevtsov2022identification} have produced domain-specific detection frameworks grounded in real campaign data.

Several studies highlight that strong detection is achievable without relying on social graph data. BiLGRU demonstrates that linguistic embeddings and textual content alone can yield robust performance \cite{wei2023twitter}, while BotShape extracts time-series characteristics from event logs to identify posting frequencies indicative of automation that static feature analysis would overlook \cite{wu2023botshape}. Human-AI hybrid ensembles have also proven effective, with combined systems outperforming either human annotators or automated models alone \cite{lagatta2026human}. At the population level, BotPercent provides a pipeline for estimating bot prevalence within communities using confidence calibration across multiple models \cite{tan2023botpercent}. Collectively, these findings underscore the fundamental difficulty of bot detection noted by Beatson et al. (2021) \cite{Beatson2021}, and motivate approaches that combine multiple complementary signals rather than relying on any single modality.

\section{Methodology}
\begin{figure}[htbp]
\centering
\includegraphics[width=.98\columnwidth]{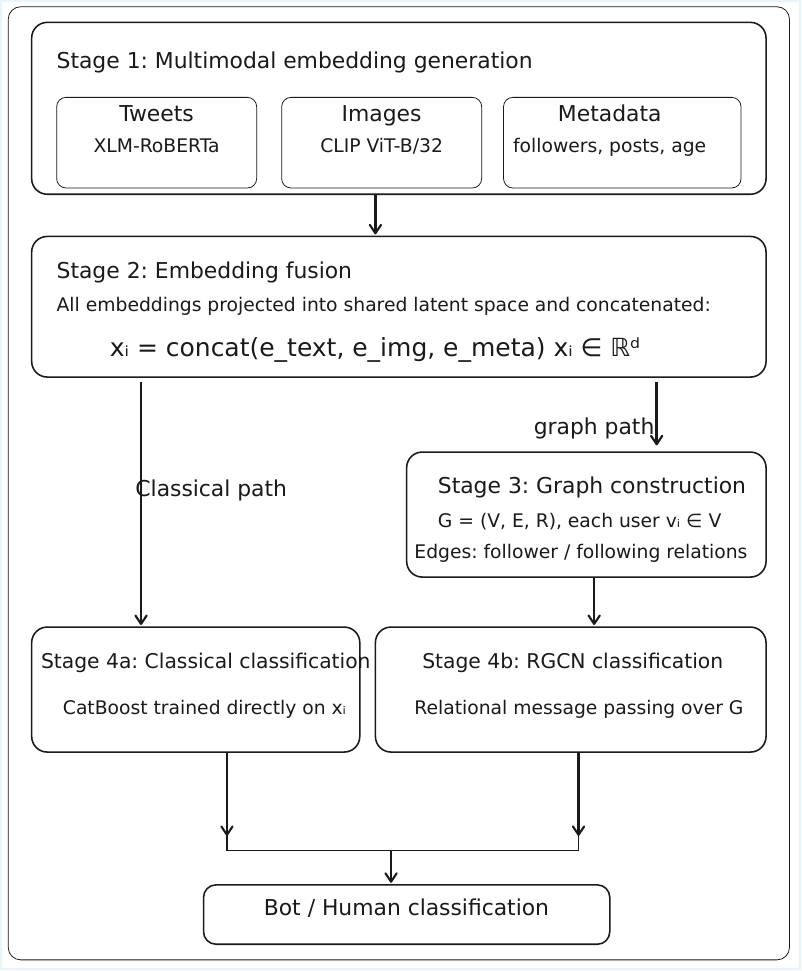}
\caption{Overview of the proposed multimodal detection framework. 
User profiles are encoded into textual (XLM-RoBERTa), visual (CLIP), 
and metadata representations, which are fused via PCA projection and 
averaging. The resulting representation is passed to two parallel 
modeling paths: a CatBoost classifier that treats users independently, 
and an R-GCN that propagates information across the social graph.}
\label{fig:architecture}
\end{figure}

This study proposes a multimodal framework for social bot detection that integrates textual, visual, and metadata signals within both classical and graph-based learning paradigms. The overall architecture of the proposed system is illustrated in Figure~\ref{fig:architecture}. Given a set of users, the framework transforms heterogeneous inputs into a unified representation, which is subsequently processed through two parallel modeling paths: a classical feature-based classifier and a graph-based relational model.

Formally, let \(\mathcal{U} = \{u_1, \dots, u_N\}\) denote a set of users, where each user \(u_i\) is associated with multimodal observations and a binary label \(y_i \in \{0,1\}\) indicating whether the account is human or bot.

\subsection{Preprocessing}
Textual data is normalized through lowercasing and tokenization, and a fixed number \(T\) of recent tweets (200 in our implementation) is selected per user to ensure uniform input size. Profile images are resized to \(224 \times 224\) pixels and normalized to the requirements of the CLIP model. Metadata features are log-transformed where necessary to reduce skewness and standardized to zero mean and unit variance. In cases where modalities are missing, zero-imputation is applied after normalization to preserve consistent feature dimensionality across all users.

\subsection{Multimodal Embedding Generation (Stage 1)}
As depicted in Stage 1 of Figure~\ref{fig:architecture}, each modality is independently encoded into a dense vector representation.

For a given user \(u_i\), textual information is processed using XLM-RoBERTa, which produces contextualized token embeddings. These are aggregated using mean pooling across tokens and then averaged across the selected tweets, resulting in a fixed-dimensional representation. This embedding captures semantic and stylistic patterns in user-generated content.

Visual information is encoded using CLIP (ViT-B/32), which maps the profile image to a vector. This representation captures high-level visual semantics and is effective in identifying patterns such as image reuse or synthetic imagery. Metadata features are also combined into a numerical vector, representing behavioral and structural properties of the account.

Both XLM-RoBERTa and CLIP encoders are used in inference mode with frozen weights; no task-specific fine-tuning is applied, as the scale of TwiBot-22 and computational constraints make full fine-tuning impractical.

\subsection{Embedding Fusion (Stage 2)}
In Stage 2 of Figure~\ref{fig:architecture}, modality-specific embeddings are transformed and combined into a unified representation suitable for downstream learning.

Let \(\mathbf{e}_{\text{text}}^{(i)} \in \mathbb{R}^{d_t}\) and \(\mathbf{e}_{\text{img}}^{(i)} \in \mathbb{R}^{d_v}\) denote the raw text and image embeddings for user \(u_i\). Since these embeddings originate from different models and lie in different feature spaces, they are first projected into a common latent space.

Specifically, Principal Component Analysis (PCA) is applied independently to each modality using the training data, producing lower-dimensional representations:

\[
\tilde{\mathbf{e}}_{\text{text}}^{(i)} \in \mathbb{R}^{d_p}, \quad
\tilde{\mathbf{e}}_{\text{img}}^{(i)} \in \mathbb{R}^{d_p}
\]

where \(d_p\) is a fixed target dimensionality (256 in our implementation).

Following projection, both embeddings are standardized using a shared scaler fit exclusively on the training split and applied to validation and test data without refitting, ensuring no leakage of distributional information across splits. Rather than concatenating high-dimensional embeddings directly, we compute a fused multimodal embedding by averaging the projected representations:

\[
\mathbf{e}_{\text{fusion}}^{(i)} = \frac{1}{2} \left( \tilde{\mathbf{e}}_{\text{text}}^{(i)} + \tilde{\mathbf{e}}_{\text{img}}^{(i)} \right)
\]

This design choice reduces redundancy and enforces alignment between modalities within the shared latent space, while maintaining a compact representation. Although learned attention-based fusion could in principle assign adaptive weights to each modality, averaging provides a parameter-free baseline that avoids overfitting on the relatively small number of users with complete multimodal observations. The contribution of each modality is instead assessed empirically through the ablation studies.

The final user representation is obtained by concatenating this fused embedding with the normalized numerical (metadata) features:

\[
\mathbf{x}_i = \text{concat}\left(\mathbf{e}_{\text{meta}}^{(i)}, \mathbf{e}_{\text{fusion}}^{(i)}\right) \in \mathbb{R}^{d_m + d_p}
\]

where \(\mathbf{e}_{\text{meta}}^{(i)} \in \mathbb{R}^{d_m}\) denotes the metadata feature vector.

To support ablation and modality-specific analysis, additional feature combinations are also constructed, including numerical-only, numerical + text, numerical + image, and text + image representations. Missing embeddings are handled via zero vectors prior to projection, ensuring consistent dimensionality across all users. While zeros in the original embedding space carry no semantic meaning, applying imputation before PCA projection means the missing signal is mapped to the origin of the learned latent space; a neutral position relative to the training distribution. This approach follows precedent in multimodal learning where modality dropout is used to encourage robustness to absent inputs, though we acknowledge that zero imputation remains an approximation and may underestimate the true impact of missing modalities.

This fusion strategy balances representational richness with computational efficiency while mitigating scale and distribution mismatches between modalities.

\subsection{Dual-Path Modeling Framework}
Following fusion, the architecture branches into two parallel paths, as illustrated in Stages 3 and 4 of Figure~\ref{fig:architecture}. Rather than constituting a controlled ablation, this dual-path design is intended as an exploratory comparison between two fundamentally different modeling paradigms: one that treats users as independent instances, and one that explicitly models their social relationships. Since the two paths differ in both architecture and access to relational information, performance differences should be interpreted as reflecting the combined effect of these factors rather than the isolated contribution of graph structure alone.

\subsubsection{Classical Path (Stage 4a):}
In the classical path, each user is treated independently. The fused representation \(\mathbf{x}_i\) is directly used as input to a CatBoost classifier, which outputs a predicted probability:

\[
\hat{y}_i = f_{\text{CatBoost}}(\mathbf{x}_i)
\]

where \(\hat{y}_i \in [0,1]\). CatBoost is particularly suitable in this setting due to its ability to model nonlinear feature interactions and handle heterogeneous data distributions effectively.

\subsubsection{Graph-Based Path (Stages 3 and 4b):}

In the graph-based path, user relationships are explicitly modeled. A directed graph \(G = (V, E, R)\) is constructed, where each node corresponds to a user and edges represent follower–following relationships. Relation types encode edge directionality, allowing the model to distinguish between incoming and outgoing connections.

Each node \(v_i \in V\) is associated with the fused feature vector \(\mathbf{x}_i\). Before relational learning, these features are projected into a latent space:

\[
\mathbf{h}_i^{(0)} = \phi(\mathbf{x}_i)
\]

where \(\phi(\cdot)\) denotes a learnable transformation consisting of a linear layer followed by layer normalization, a ReLU activation, and dropout. This step ensures that input features are mapped to a representation suitable for graph-based propagation.

Node representations are then updated using stacked Relational Graph Convolutional Network (R-GCN) layers. The relational message passing at layer \(l\) is defined as:

\[
\tilde{\mathbf{h}}_i^{(l+1)} =
\mathbf{W}_0^{(l)} \mathbf{h}_i^{(l)} +
\sum_{r \in R}
\sum_{j \in \mathcal{N}_r(i)}
\frac{1}{c_{i,r}}
\mathbf{W}_r^{(l)} \mathbf{h}_j^{(l)}
\]

where \(\mathcal{N}_r(i)\) denotes the set of neighbors of node \(i\) under relation \(r\), \(\mathbf{W}_0^{(l)}\) is a relation-independent weight matrix applied to the node's own representation, \(\mathbf{W}_r^{(l)}\) are relation-specific learnable weight matrices, and \(c_{i,r}\) is a normalization constant, typically the number of neighbors under relation \(r\).

The aggregated representation is then normalized and passed through a nonlinear transformation. To improve training stability and mitigate over-smoothing, a residual connection is applied:

\[
\mathbf{h}_i^{(l+1)} =
\text{ReLU}(\text{LayerNorm}(\tilde{\mathbf{h}}_i^{(l+1)})) + \mathbf{h}_i^{(l)}
\]

This iterative process enables each node to incorporate information from its local neighborhood, allowing the model to capture structural dependencies and coordinated behaviors that are not observable from individual features alone.

After \(L\) layers, the final node representation \(\mathbf{h}_i^{(L)}\) is passed through a multi-layer perceptron:

\[
\mathbf{z}_i = f_{\text{MLP}}(\mathbf{h}_i^{(L)})
\]

where \(\mathbf{z}_i \in \mathbb{R}^2\) represents the logits for the two classes (bot and human). The predicted probability \(\hat{y}_i\) is obtained by applying the softmax function over \(\mathbf{z}_i\).

\section{Experiments}

\subsection{Dataset}
We conduct experiments on the TwiBot-22 dataset, which provides predefined training, validation, and test splits. However, due to severe class imbalance in the original data, in which human accounts constitute 86\% of 1,000,000 users in TwiBot-22, a balanced subsample of 30,000 users is constructed for our experiments.

Specifically, we randomly sample an equal number of human and bot accounts, resulting in a dataset of 30,000 users. To ensure multimodal consistency, we retain only users with accessible profile images and complete metadata. The resulting dataset is partitioned into training, validation, and test sets following an 70/20/10 split.

For the graph-based model, the social network is constructed using follower–following relationships among the sampled users. As a result of sub-sampling, the graph is significantly sparser than the full TwiBot-22 network, which impacts the availability of relational information.

All models are trained using the training split, with hyperparameters selected based on validation performance. Final results are reported on the held-out test set.

\subsection{Evaluation Protocol}
Both the classical and graph-based modeling paths produce probabilistic predictions for each user, which are evaluated using a comprehensive set of classification metrics.

Given predicted probabilities \(\hat{y}_i\) and ground truth labels \(y_i \in \{0,1\}\), predictions are converted into binary decisions using a threshold selected on the validation set. The evaluation focuses on both threshold-dependent and threshold-independent metrics to provide a complete assessment of model performance.

Accuracy is used to measure the overall proportion of correctly classified instances. However, since bot detection often involves class imbalance, we place particular emphasis on the F1-score, which balances precision and recall and provides a more informative measure of performance on the minority class.

To evaluate model robustness across different decision thresholds, we report the Receiver Operating Characteristic Area Under the Curve (ROC-AUC), which captures the trade-off between true positive rate and false positive rate. In addition, we include the Precision–Recall Area Under the Curve (PR-AUC), which is especially informative in imbalanced settings, as it focuses on performance with respect to the positive (bot) class.

Finally, to assess the contribution of each modality, we conduct ablation studies by systematically removing text, image, or metadata features. The classical and graph-based paths are compared exploratorily to characterize the relative utility of instance-level versus relational modeling, with the caveat that architectural differences between the two paths preclude attribution of performance differences to graph structure alone. This allows us to quantify the impact of multimodal fusion and relational learning on overall performance.

To determine whether observed differences in performance are statistically meaningful, we perform significance testing on model predictions over the shared test set. Specifically, we first apply \textbf{Cochran's Q test} as an omnibus test to assess whether there are overall differences among multiple models evaluated on the same samples. We also conduct \textbf{pairwise McNemar's tests} between model pairs to identify where the differences occur. A Bonferroni correction is applied to control for multiple comparisons, ensuring a conservative significance threshold.

The impact of frozen versus fine-tuned encoders is not evaluated in this work and represents a direction for future investigation.

\subsection{Baselines}
To contextualize the performance of the proposed models, we compare against two reference points.

\subsubsection{Numerical-only model:}
The CatBoost classifier trained exclusively on metadata features, with no textual or visual signals. This serves as the primary internal baseline, isolating the contribution of multimodal fusion by measuring how much additional discriminative power text and image embeddings provide beyond account-level statistics alone.

\subsubsection{Published TwiBot-22 Results:}
We reference performance reported on BotRGCN \cite{feng2021botrgcn} and RGT \cite{feng2022rgt}, both of which were evaluated on the full TwiBot-22 benchmark \cite{twibot-22}. BotRGCN reports an accuracy of 0.797 and an F1 score of 0.575, while RGT reports an accuracy of 0.765 and an F1 score of 0.429. These figures provide an upper reference point for what graph-based models can achieve on this dataset under full-graph conditions, and highlight that F1 scores on TwiBot-22 are substantially lower than on TwiBot-20, reflecting the greater difficulty of the benchmark.

It should be noted that direct numerical comparison with published results is approximate. Our evaluation is conducted on a balanced subsample of 30,000 users rather than the entire dataset, and the sampled social graph is substantially sparser than the complete TwiBot-22 network. Performance differences should therefore be interpreted in light of these differing experimental conditions rather than as a direct measure of relative model quality.

\section{Experimental Results}
This section presents the empirical evaluation of the proposed multimodal framework. We first report the performance of the classical (CatBoost-based) models across different modality combinations, followed by the graph-based model, and situate both against published baselines on TwiBot-22. Statistical significance testing is then used to assess whether observed differences are meaningful.

\subsection{Overall Performance}
Table~\ref{tab:main_results} summarizes the performance of all models on the test set. The results include both threshold-dependent metrics (Accuracy, F1) and threshold-independent metrics (ROC-AUC, PR-AUC), providing a comprehensive evaluation.

\begin{table}[htbp]
\centering
\resizebox{.98\columnwidth}{!}{
\begin{tabular}{lcccc}
\textbf{Model} & \textbf{Accuracy} & \textbf{F1 (Weighted)} 
    & \textbf{ROC-AUC} & \textbf{PR-AUC} \\[0.1cm]
\hline
BotRGCN \cite{feng2021botrgcn}$^\dagger$ & 0.797 & 0.575 & -- & -- \\
RGT \cite{feng2022rgt}$^\dagger$         & 0.765 & 0.429 & -- & -- \\
\hline
Num (internal baseline)  & 0.716 & 0.721 & 0.756 & 0.841 \\
Num + Image              & 0.719 & 0.722 & \textbf{0.757} & \textbf{0.842} \\
Num + Text               & 0.706 & 0.711 & 0.751 & 0.841 \\
Num + Text + Image       & \textbf{0.720} & \textbf{0.724} & 0.756 & 0.839 \\
Graph (R-GCN)            & 0.691 & 0.698 & 0.739 & 0.602 \\
\hline
\end{tabular}
}
\caption{Test set performance across feature configurations and modeling approaches. Best results among our models are shown in bold. $^\dagger$Published results on the \textit{full} TwiBot-22 dataset; ROC-AUC and PR-AUC are not reported in the original works.}
\label{tab:main_results}
\end{table}

The best overall performance among our models is achieved by the Num + Text + Image configuration, with an accuracy of 0.720 and a weighted F1 score of 0.724. Notably, this F1 score exceeds those reported by both BotRGCN (0.575) and RGT (0.429) on TwiBot-22, despite operating on a balanced subsample with a fragmented social graph. This result should be interpreted cautiously given the differing experimental conditions, but it suggests that metadata-driven multimodal fusion can be competitive with graph-based methods on this benchmark, particularly when full graph access is unavailable.

The improvement over the numerical-only baseline is marginal across all multimodal variants, indicating that metadata features alone provide a strong 
predictive signal. While the inclusion of image features slightly improves 
ROC-AUC and PR-AUC, these gains remain limited.

\subsection{Impact of Multimodal Fusion}
The results demonstrate that naïve multimodal fusion provides only incremental improvements over the numerical baseline. Adding image features leads to small but consistent gains in ranking metrics, while text features do not significantly improve performance and, in some cases, slightly degrade it.

The combined multimodal representation (Num+Text+Image) yields the most stable performance across metrics, suggesting that averaging projected embeddings before concatenation helps reduce redundancy. Nevertheless, the overall gains remain modest, indicating that the additional modalities do not substantially enhance discriminative power beyond metadata features.

A class-wise analysis further reveals an imbalance in performance. All classical models achieve strong detection of bots (\(F1 \approx 0.78\)), but performance on human accounts remains lower (\(F1 \approx 0.60\)), indicating a tendency toward false positives when classifying legitimate users.

\subsection{Graph-Based Model Performance}
The graph-based model achieves an accuracy of 0.691 and a weighted F1 score of 0.698, underperforming all classical variants. Its ROC-AUC (0.739) is also lower, and its PR-AUC (0.602) drops substantially, indicating weaker performance in identifying the positive class.

In terms of F1, however, our R-GCN (0.698) still exceeds the published F1 scores of BotRGCN (0.575) and RGT (0.429) on TwiBot-22. This is consistent with the interpretation that F1 scores on TwiBot-22 are generally lower than on TwiBot-20, reflecting the greater difficulty of the benchmark, and that our balanced sampling may produce favorable F1 conditions relative to the full imbalanced dataset.

Despite lower overall accuracy, the graph model exhibits a more balanced treatment of classes, with improved bot recall and stronger human classification performance relative to earlier training stages. This suggests that relational modeling captures complementary structural information, but not at a level sufficient to outperform feature-based approaches under the fragmented graph conditions of this study. This interpretation is consistent with ablation evidence reported by \citet{twibot-22}, who show that removing graph components from BotRGCN and RGT causes accuracy drops of 4.3 and 1.1 points respectively on TwiBot-22, with F1 degradation as large as 16.3 points. Our sub-sampled graph conditions are analogous to these degraded settings, suggesting that the performance gap reflects structural data limitations rather than a fundamental weakness of relational modeling.

\subsection{Statistical Significance Analysis}
To assess whether observed differences are statistically significant, we performed significance testing on predictions over the shared test set.

\begin{table}[htbp]
\centering
\resizebox{.98\columnwidth}{!}{
\begin{tabular}{lccccc}
\textbf{Model} & \textbf{Num} & \textbf{Num+Img} & \textbf{Num+Text} 
    & \textbf{Num+Text+Img} & \textbf{Graph} \\[0.1cm]
\hline
Num                & --   & ns   & ns   & ns            & \textbf{s} \\
Num+Img            & ns   & --   & ns   & ns            & \textbf{s} \\
Num+Text           & ns   & ns   & --   & \textbf{s}    & \textbf{s} \\
Num+Text+Img       & ns   & ns   & \textbf{s} & --       & \textbf{s} \\
Graph              & \textbf{s} & \textbf{s} & \textbf{s} & \textbf{s} & -- \\
\hline
\end{tabular}
}
\caption{Pairwise statistical significance (McNemar's test with Bonferroni correction). \textbf{'s'} indicates a statistically significant difference (\(p < 0.05\) after correction); \textbf{'ns'} denotes non-significant differences.}
\label{tab:significance}
\end{table}

Cochran's Q test reveals a significant overall difference across models (Q = 2046.95, \(p < 0.001\)). Pairwise McNemar's tests with Bonferroni correction show that differences among classical models are generally not statistically significant despite small numerical variations, a significant difference is observed between Num + Text and Num + Text + Image indicating a measurable contribution from image features when combined with text, and all comparisons between the graph-based and classical models are statistically significant, confirming that the observed performance gap is robust. These findings reinforce that while multimodal fusion introduces slight improvements, they are not consistently significant, whereas the difference between modeling paradigms is substantial.

\section{Discussion}
\subsection{Revisiting the Research Questions}

\textbf{RQ1: Does combining visual and textual features improve bot detection?} The results provide a nuanced answer. While the full multimodal configuration (Num + Text + Image) achieves the highest accuracy and weighted F1 among our models, the gains over the numerical-only baseline are marginal and largely not statistically significant. This suggests that metadata features such as follower counts, account age, and posting activity already encode much of the discriminative signal available in the dataset. Visual and textual embeddings contribute incrementally but do not fundamentally alter the decision boundary learned by the classifier.

One plausible explanation is that the frozen encoder strategy limits the expressiveness of the extracted representations. Since XLM-RoBERTa and CLIP are used in inference mode without task-specific fine-tuning, the resulting embeddings may not capture the subtle stylistic and visual cues that distinguish bot-generated content from human-generated content. Fine-tuning on bot detection data could potentially yield stronger multimodal gains, and represents a natural direction for future work.

\textbf{RQ2: Can social network graphs improve performance on the bot detection task?} In this study, the graph-based R-GCN underperforms all classical variants across accuracy, F1, and ROC-AUC, with a particularly sharp drop in PR-AUC. However, this result should not be interpreted as evidence against graph-based modeling in general. Rather, it reflects the structural limitations imposed by sub-sampling. Restricting the analysis to 30,000 users fragments the original follower–following network, severing connections to users outside the sampled set and producing a sparser graph. This is consistent with ablation findings from \citet{twibot-22}, where removing graph components from BotRGCN and RGT produces similarly degraded results on TwiBot-22.

Notably, the performance difference between the graph model and all classical variants is statistically significant, confirming that the two paradigms produce meaningfully different predictions. The graph model's more balanced class treatment suggests that structural signals capture complementary information even under degraded graph conditions, pointing to potential value in hybrid approaches that combine relational and feature-based signals.

\subsection{Comparison with Published Results}
Our classical models achieve F1 scores between 0.711 and 0.724, exceeding the F1 scores reported by BotRGCN (0.575) and RGT (0.429) on TwiBot-22. This comparison should be interpreted carefully: our evaluation uses a balanced subsample rather than the full dataset, which inflates F1 relative to imbalanced evaluation conditions. Nevertheless, the result demonstrates that metadata-driven multimodal fusion can be competitive with graph-based approaches on this benchmark, particularly when full social graph access is unavailable or computationally prohibitive.

\subsection{The Role of Metadata}
A consistent finding across all experiments is the strong predictive power of numerical metadata features. Account-level statistics such as follower-to-following ratios, tweet frequency, and account age provide robust signals that prove difficult for adversaries to simultaneously manipulate at scale. This aligns with prior work showing that behavioral and structural metadata remain among the most reliable indicators of automation \cite{rossetti_bots_2023, feng2021botrgcn}.

The limited incremental benefit of adding text and image features may also reflect the nature of TwiBot-22, which was constructed to include sophisticated bots capable of generating plausible content. In such settings, textual and visual signals are naturally harder to distinguish from those of genuine users, placing greater relative weight on metadata.

\subsection{Class Imbalance in Predictions}
Despite balanced sampling during data preparation, all classical models exhibit asymmetric class performance, achieving strong bot detection (F1~\(\approx\)~0.78) but weaker human classification (F1~\(\approx\)~0.60). This asymmetry suggests that the learned decision boundary is biased toward the bot class, likely due to the nature of the feature representations rather than label imbalance. Future work could explore calibration techniques or asymmetric loss functions to improve performance on legitimate user classification, which is critical for minimizing false positives in deployed systems.

\subsection{Limitations}
Several limitations bound the conclusions of this study. First, the use of frozen encoders constrains the expressiveness of visual and textual representations. Second, sub-sampling for class balance and image reachability filters the dataset in ways that may not reflect real-world distributions. Third, graph sub-sampling likely underestimates the true benefit of relational modeling. Finally, the dual-path comparison conflates architectural differences with access to relational information, making it impossible to isolate the contribution of graph structure alone.

\subsection{Implications}
Despite these limitations, this study offers two practical insights. First, metadata-based detection remains a strong and computationally efficient baseline that should not be overlooked in favor of more complex multimodal pipelines. Second, graph-based modeling shows promise for capturing coordination patterns but requires access to a sufficiently dense network to be effective. Practitioners deploying bot detection systems should consider the trade-off between social graph completeness and the overhead of relational modeling.

\section{Conclusion and Future Work}
This research presented a multimodal framework for social bot detection on Twitter, integrating textual, visual, and metadata signals within both classical and graph-based learning paradigms. Evaluated on a balanced 
subsample of the TwiBot-22 benchmark, our results demonstrate that metadata features provide a strong and stable predictive signal, with multimodal fusion offering incremental but largely non-significant improvements under frozen encoder conditions. The full multimodal configuration (Num + Text + Image) achieves the strongest overall performance among our models, with F1 scores that compare favorably to published graph-based methods on TwiBot-22 despite the absence of full graph access. The graph-based R-GCN underperforms classical variants, a result we attribute to graph fragmentation caused by sub-sampling rather than to a fundamental limitation of relational modeling. Statistical significance testing confirms that the performance gap between modeling paradigms is robust, while differences among classical multimodal configurations are generally not significant, suggesting that feature quality matters more than fusion strategy under the conditions studied here.

Several directions emerge directly from these findings. The most immediate is \textbf{fine-tuning pretrained encoders}: replacing frozen XLM-RoBERTa and CLIP with task-adapted variants could yield richer representations that better capture the stylistic and visual cues characteristic of automated accounts, potentially unlocking the multimodal gains that averaged fusion was unable to produce in this work. Complementing this, replacing the averaging-based fusion with \textbf{learned attention or gating mechanisms} would allow the model to dynamically weight modalities based on their reliability for a given user, improving robustness when modalities are missing or noisy.

On the graph side, future work should evaluate the R-GCN on a denser subgraph using scalable sampling strategies such as GraphSAINT or Cluster-GCN that preserve neighborhood structure while managing memory constraints. Combining relational and feature-based signals within a \textbf{hybrid classical-graph model} represents a particularly promising direction, given that the two paradigms produce statistically distinct predictions and likely capture complementary information. Beyond architecture, incorporating \textbf{temporal signals} from posting histories, as explored in BotShape \cite{wu2023botshape} and BotDGT \cite{he2024botdgt}, could improve detection of bots that mimic human behavior at any given moment but reveal automation through longitudinal patterns.

Finally, evaluating the framework across platforms such as Reddit or Instagram would test generalization beyond Twitter's specific behavioral norms, while probing \textbf{adversarial robustness} against bots that strategically manipulate metadata or generate synthetic imagery would assess real-world deployability. As automated accounts grow more sophisticated, addressing these directions within a unified and scalable framework remains an important open challenge for the field.

\bibliography{aaai2026}

@article{x_fake,
  author = {Dracewicz, Weronika and Sepczuk, Mariusz},
  title = {Detecting Fake Accounts on Social Media Portals—The X Portal Case Study},
  journal = {Electronics},
  volume = {13},
  year = {2024},
  number = {13},
  doi = {10.3390/electronics13132542}
}

@article{sallah2024finetuned,
  title={Fine-Tuned Understanding: Enhancing Social Bot Detection with Transformer-based Classification},
  author={Sallah, Amine and others},
  journal={IEEE Access},
  volume={12},
  pages={106308--106325},
  year={2024},
  doi={10.1109/ACCESS.2024.3440657}
}

@misc{twibot-22,
  author = {Feng, Shangbin and others},
  title = {TwiBot-22: Towards Graph-Based Twitter Bot Detection},
  publisher = {arXiv},
  year = {2022},
  doi = {10.48550/ARXIV.2206.04564}
}

@inproceedings{feng2021botrgcn,
  title={Bot-RGCN: Twitter Bot Detection with Relational Graph Convolutional Networks},
  author={Feng, Shangbin and Wan, Herun and Wang, Ningnan and Li, Jundong and Luo, Minnan},
  booktitle={2021 IEEE/ACM International Conference on Advances in Social Networks Analysis and Mining (ASONAM)},
  year={2021}
}

@inproceedings{feng2022rgt,
  title={Heterogeneity-Aware Twitter Bot Detection with Relational Graph Transformers},
  author={Feng, Shangbin and Tan, Zhaoxuan and Li, Rui and Luo, Minnan},
  booktitle={Proceedings of the 36th AAAI Conference on Artificial Intelligence (AAAI)},
  year={2022}
}

@article{anshul2025rogbot,
  title={RoGBot: Relationship-Oblivious Graph-based Neural Network with Contextual Knowledge for Bot Detection},
  author={Anshul, Ashutosh and Rehman, Mohammad Zia Ur and Kadali, Sri Akash and Kumar, Nagendra},
  journal={arXiv preprint arXiv:2510.23648},
  year={2025}
}

@article{wu2025msmbd,
  title={MSM-BD: Multimodal Social Media Bot Detection Using Heterogeneous Information},
  author={Wu, Tingxuan and Ma, Zhaorui and Cui, Yanjun and Zhou, Ziyi and Wang, Eric},
  journal={arXiv preprint arXiv:2501.00204},
  year={2025}
}

@article{cheng2025cbmte,
  title={CB-MTE: Social Bot Detection via Multi-Source Heterogeneous Feature Fusion},
  author={Cheng, Meng and Xiao, Yuzhi and Huang, Tao and Lei, Chao and Zhang, Chuang},
  journal={Sensors},
  volume={25},
  number={11},
  pages={3549},
  year={2025}
}

@article{wu2023botshape,
  title={BotShape: A Novel Social Bots Detection Approach via Behavioral Patterns},
  author={Wu, Jun and Ye, Xuesong and Mou, Chengjie},
  journal={arXiv preprint arXiv:2303.10214},
  year={2023}
}

@inproceedings{tan2023botpercent,
  title={BotPercent: Estimating Bot Populations in Twitter Communities},
  author={Tan, Zhaoxuan and Feng, Shangbin and Sclar, Melanie and Wan, Herun and Luo, Minnan and Choi, Yejin and Tsvetkov, Yulia},
  booktitle={Findings of the Association for Computational Linguistics: EMNLP 2023},
  year={2023}
}

@article{huang2025srgat,
  title={Semi-Supervised Social Bot Detection with Relational Graph Attention Transformers},
  author={Huang, Di and Song, Jinbao and Zhang, Xingyu},
  journal={Information Fusion},
  volume={118},
  year={2025}
}

@inproceedings{he2024botdgt,
  title={BotDGT: Dynamicity-aware Social Bot Detection with Dynamic Graph Transformers},
  author={He, Buyun and others},
  booktitle={Proceedings of the IJCAI},
  year={2024}
}

@inproceedings{zhang2024sega,
  title={SeGA: Preference-Aware Self-Contrastive Learning with Prompts for Anomalous User Detection},
  author={Zhang, Ying and others},
  booktitle={Proceedings of the 38th AAAI Conference},
  year={2024}
}

@inproceedings{feng2024whatbot,
  title={What Does the Bot Say? Opportunities and Risks of LLMs in Social Media Bot Detection},
  author={Feng, Shangbin and others},
  booktitle={Proceedings of the 62nd Annual Meeting of the ACL},
  year={2024}
}

@inproceedings{liu2023botmoe,
  title={BotMoE: Twitter Bot Detection with Community-Aware Mixtures of Experts},
  author={Liu, Yuhan and others},
  booktitle={Proceedings of the 46th International ACM SIGIR},
  year={2023}
}

@article{ArranzEscudero2025EnhancingMCA,
  title={Enhancing misinformation countermeasures: a multimodal approach to twitter bot detection},
  author={Arranz-Escudero, O. and others},
  journal={Social Network Analysis and Mining},
  year={2025},
  volume={15}
}

@article{Beatson2021,
  author = {Beatson, O. and others},
  title ={Automation on Twitter: Measuring the Effectiveness of Approaches to Bot Detection},
  journal = {Social Science Computer Review},
  volume = {41},
  year = {2023}
}

@article{choudhary_mmds_2024,
  author = {Choudhary, Monika and Patil, Spandan and Chouhan, Satyendra Singh and Pilli, Emmanuel S.},
  title = {MmDs: Multimodal benchmark dataset for suspicious profile detection on twitter social network},
  journal = {Social Network Analysis and Mining},
  volume = {14},
  number = {1},
  pages = {129},
  year = {2024},
  doi = {10.1007/s13278-024-01296-9}
}

@article{sharma2023fakedbits,
  title={FakedBits- Detecting Fake Information on Social Platforms using Multi-Modal Features},
  author={Sharma, Dilip Kumar and Singh, Bhuvanesh and Agarwal, Saurabh and Kim, Hyunsung and Sharma, Raj},
  journal={KSII Transactions on Internet and Information Systems},
  volume={17},
  number={1},
  pages={51--71},
  year={2023},
  doi={10.3837/tiis.2023.01.004}
}

@inproceedings{zhou2023cbd,
  title={Detecting Social Bot on the Fly using Contrastive Learning},
  author={Zhou, Ming and Zhang, Dan and Wang, Yuandong and Geng, Yangli-ao and Tang, Jie},
  booktitle={Proceedings of the 32nd ACM International Conference on Information and Knowledge Management (CIKM '23)},
  pages={4995--5001},
  year={2023},
  doi={10.1145/3583780.3615468}
}

@article{bruno2022brexit,
  title={Brexit and bots: characterizing the behaviour of automated accounts on Twitter during the UK election},
  author={Bruno, Matteo and Lambiotte, Renaud and Saracco, Fabio},
  journal={EPJ Data Science},
  volume={11},
  number={1},
  pages={1--26},
  year={2022},
  doi={10.1140/epjds/s13688-022-00330-0}
}

@inproceedings{shevtsov2022identification,
  title={Identification of Twitter Bots Based on an Explainable Machine Learning Framework: The US 2020 Elections Case Study},
  author={Shevtsov, Alexander and Tzagkarakis, Christos and Antonakaki, Despoina and Ioannidis, Sotiris},
  booktitle={Proceedings of the Sixteenth International AAAI Conference on Web and Social Media (ICWSM 2022)},
  pages={939--950},
  year={2022}
}

@article{gera2022tbot,
  title={T-Bot: AI-based social media bot detection model for trend-centric twitter network},
  author={Gera, Suruchi and Sinha, Adwitiya},
  journal={Social Network Analysis and Mining},
  volume={12},
  number={1},
  pages={76},
  year={2022},
  doi={10.1007/s13278-022-00897-6}
}

@article{wan2026hcgbot,
  title={HCGBot: Learning Homophilous Context Graphs for Twitter Bot Detection},
  author={Wan, Herun and Luo, Minnan and Wang, Jihong and Chang, Xiaojun and Zheng, Qinghua},
  journal={IEEE Transactions on Knowledge and Data Engineering},
  volume={38},
  number={3},
  pages={1813--1827},
  year={2026},
  doi={10.1109/TKDE.2026.3656720}
}

@article{lagatta2026human,
  title={Human, AI, and Hybrid Ensembles for Detection of Adaptive, RL-based Social Bots},
  author={La Gatta, Valerio and Subrahmanian, Nathan and Wang, Kaitlyn and Birnbaum, Larry and Subrahmanian, V.S.},
  journal={arXiv preprint arXiv:2603.23796},
  year={2026}
}

@article{wang2025hhgbot,
  title={HHG-Bot: A Hyperheterogeneous Graph-Based Twitter Bot Detection Model},
  author={Wang, Tianbo and Wang, Zhao and Li, Huacheng and Xia, Chunhe and Zhao, Chun},
  journal={IEEE Transactions on Computational Social Systems},
  volume={12},
  number={5},
  pages={3416--3430},
  year={2025},
  doi={10.1109/TCSS.2025.3543419}
}

@article{ng2025global,
  title={A global comparison of social media bot and human characteristics},
  author={Ng, Lynnette Hui Xian and Carley, Kathleen M.},
  journal={Scientific Reports},
  volume={15},
  number={1},
  pages={10973},
  year={2025},
  doi={10.1038/s41598-025-96372-1}
}

@inproceedings{bui2022twitter,
  title={Twitter Bot Detection using Social Network Analysis},
  author={Bui, Thi and Potika, Katerina},
  booktitle={2022 Fourth International Conference on Transdisciplinary AI (TransAI)},
  pages={55--58},
  year={2022},
  doi={10.1109/TransAI54797.2022.00022}
}

@article{wei2023twitter,
  title={Twitter Bot Detection Using Neural Networks and Linguistic Embeddings},
  author={Wei, Feng and Nguyen, Uyen Trang},
  journal={IEEE Open Journal of the Computer Society},
  volume={4},
  pages={219--231},
  year={2023},
  doi={10.1109/OJCS.2023.3302286}
}

@inproceedings{lin2025dmibot,
  title={DMIBot: Dynamic Multimodal Interaction for Twitter Bot Detection},
  author={Lin, Xiezhuo and Wu, Qingfeng and Huang, Yurui},
  booktitle={ICASSP 2025 - 2025 IEEE International Conference on Acoustics, Speech and Signal Processing (ICASSP)},
  pages={1--5},
  year={2025},
  doi={10.1109/ICASSP49660.2025.10889631}
}

@article{rossetti_bots_2023,
	title = {Bots, disinformation, and the first impeachment of U.S. President Donald Trump},
	volume = {18},
	issn = {1932-6203},
	url = {https://journals.plos.org/plosone/article?id=10.1371/journal.pone.0283971},
	doi = {10.1371/journal.pone.0283971},
	pages = {e0283971},
	number = {5},
	journal = {{PLOS} {ONE}},
	shortjournal = {{PLOS} {ONE}},
	publisher = {Public Library of Science},
	author = {Rossetti, Michael and Zaman, Tauhid},
	urldate = {2026-04-27},
	date = {2023-05-08},
    year = {2023},
	langid = {english},
}
\end{document}